\documentclass[10pt,twocolumn,letterpaper]{article}

\usepackage{iccv}
\usepackage{times}
\usepackage{epsfig}
\usepackage{graphicx}
\usepackage{amssymb}
\usepackage{booktabs}
\usepackage{amssymb}
\usepackage{float}
\usepackage{amsmath}
\usepackage{setspace}
\usepackage{multirow}

\usepackage[pagebackref=true, breaklinks=true, colorlinks, bookmarks=false]{hyperref}

\iccvfinalcopy 

\def\iccvPaperID{6092} 

\newcommand{\hupar}[1]{\vspace{12pt} \noindent \textbf{#1}\quad}

\begin{document}

\title{A robust visual sampling model inspired by receptive field}

\author{Liwen Hu\textsuperscript{1},
Lei Ma\textsuperscript{1}, 
Dawei Weng\textsuperscript{2},
Tiejun Huang\textsuperscript{1}\\
\textsuperscript{1}Department of Computer Science and Technology, Peking University\\
\textsuperscript{2}Biomedical Engineering, Capital Medical University
}

\maketitle

\begin{abstract}
 Spike camera mimicking the retina fovea can report per-pixel luminance intensity accumulation by firing spikes. As a bio-inspired vision sensor with high temporal resolution, it has a huge potential for computer vision. However, the sampling model in current Spike camera is so susceptible to quantization and noise that it cannot capture the texture details of objects effectively. 
In this work, a robust visual sampling model inspired by receptive field (RVSM) is proposed where wavelet filter generated by difference of Gaussian (DoG) and Gaussian filter are used to simulate receptive field. Using corresponding method similar to inverse wavelet transform, spike data from RVSM can be converted into images. To test the performance, we also propose a high-speed motion spike dataset (HMD) including a variety of motion scenes. By comparing reconstructed images in HMD, we find RVSM can improve the ability of capturing information of Spike camera greatly. More importantly, due to mimicking receptive field mechanism to collect regional information, RVSM can filter high intensity noise effectively and improves the problem that Spike camera is sensitive to noise largely. Besides, due to the strong generalization of sampling structure, RVSM is also suitable for other neuromorphic vision sensor. Above experiments are finished in a Spike camera simulator 
\end{abstract}
\section{Introduction}
With the development of computing power and AI (artificial intelligence), significant progress has been made in unmanned aerial vehicles, autonomous driving and visual tracking. The applications have to deal with many challenging scenarios including high-speed motion and low light condition in real time. However, common traditional cameras with frame-based paradigms are not a fine choice to capture the high-speed motion of objects due to their low time resolution. Besides, video streams from the traditional cameras are not conducive to real-time computing because the dense and redundant data brings extra computational costs to classical algorithms such as convolutional neural networks (CNNs) \cite{net1, net2, net3, net4, net5}.  
\\\indent Event cameras referring to the human visual system can asynchronously output sparse event streams according to the change of brightness and have high temporal resolution. Hence, it is more suitable for robotics and computer vision tasks in high-speed motion scene than traditional cameras. However, event cameras cannot capture the visual texture of objects such as dynamic vision sensor (DVS)  \cite{dvs1, dvs2}. Although some event cameras can solve above problems by combing DVS and conventional image sensor (DAVIS  \cite{dvs3}), or adding an extra photo-measurement circuit (ATIS  \cite{dvs4}, CeleX  \cite{dvs5}), there is motion mismatch due to the difference of the sampling time resolution between DVS and extra photo-measurement circuit.
Similar to event cameras, Spike camera \cite{spikecamera} is also inspired by biological visual systems. Specifically, Spike camera models integrate-and-fire neurons and can report per-pixel brightness accumulation by outputting a sparse spike data. Hence, it has not only the similar advantage as event cameras, i.e. high temporal resolution (40000Hz), but also it can capture the visual texture of objects. Although Spike camera can theoretically sample in all kinds of scene, its sampling model (FSM \cite{spikecamera}) in complex environments is not ideal enough due to the presence of quantization (as fig3) and noise (as fig4). And the sampling of Spike camera is susceptible to noise largely while human visual systems are robust to noise. Therefore, there still exists significant potential for the improvement of the bio-inspired sampling model.\\\indent
To this end, our aim is to improve the ability of Spike camera to capture the texture information in high-speed motion scenes. Our main contributions are summarized as follows:\\\indent
\begin{figure*}[ht]
\includegraphics[width=\linewidth]{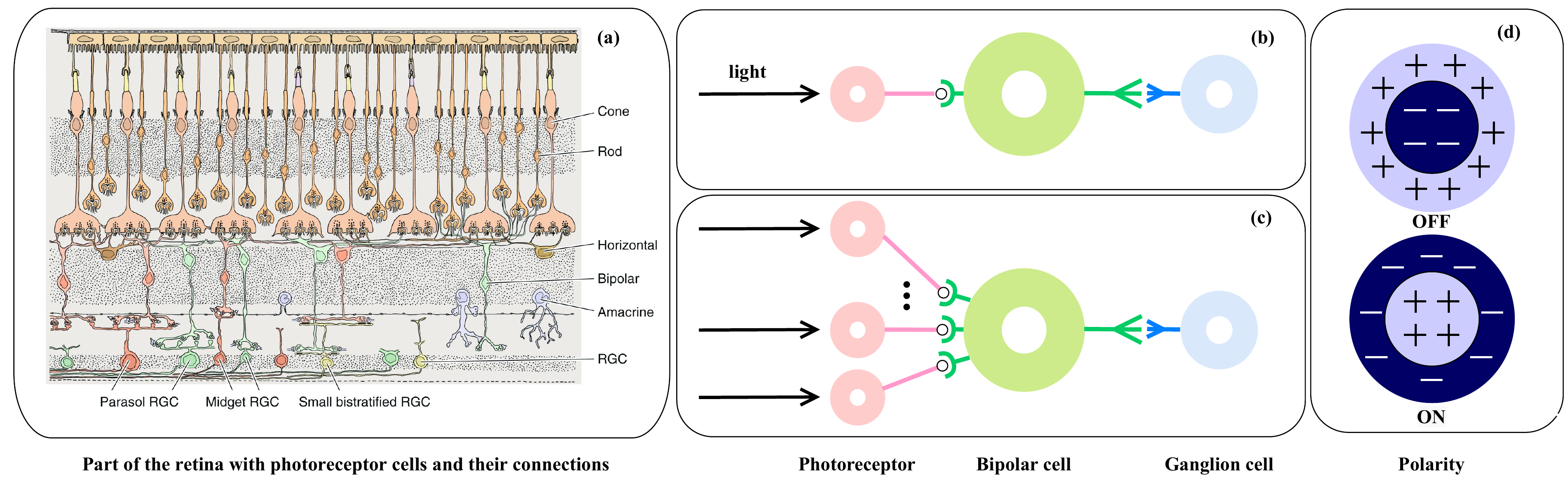}
\centering
\caption{The connection structure of cells in the human visual system. (a) denotes the actual connection structure of cells in retina \cite{bio5}. (b) denotes the simplified connection structure in FSM. (c) denotes the critical connection structure ignored by FSM. (d) shows ``Off" and ``on" receptive field. ``+" represents that corresponding position is sensitive to brightening, and ``-" represents that corresponding position is sensitive to darkening.}\label{fig1}
\end{figure*}
\begin{itemize}
\item[1)]
 We propose a novel and robust visual sampling model inspired by receptive field (RVSM) where wavelet filter bank generated by DoG (RVSM$_{Dog}$) and Gaussian filter bank (RVSM$_{Gauss}$) are used to mimic the receptive field mechanism of ganglion cells in human retina respectively.
\end{itemize}
\begin{itemize}
\item[2)]
An efficient method to convert spike data from RVSM into images is proposed which is similar to inverse wavelet transform. By comparing images reconstructed from spike data in our dataset, we find RVSM can capture much more texture details in motion scenes than FSM. Besides, by collecting regional information, RVSM can filter high intensity noise effectively which is consistent with our understanding of the human visual system.
\end{itemize}
\begin{itemize}
\item[3)]
We propose a high-speed motion spike dataset (HMD) which covers various motion scenes (single object and multi-object motion). HMD includes spike data from RVSM$_{Dog}$, RVSM$_{Gauss}$ and FSM respectively generated by Spike camera simulator and corresponding image sequences. 
\end{itemize}
\section{Related Work and Expansion}


\subsection{Fovea-like Sampling Method}
As a bio-inspired sampling method on Spike camera, FSM mainly mimics fovea in human visual system, where a photoreceptor first converts optical signal into electrical signal, then some bipolar cell processes electrical signal and sends it to some connected ganglion cell, finally the ganglion cell decides whether to output a spike \cite{fovea1, fovea2}. FSM conceives the above visual sampling process as a summation process (sometimes also called ‘integration’ process) combined with a mechanism that triggers action potentials above some critical voltage (as fig2).
\begin{figure}[htb]
\includegraphics[width=8cm]{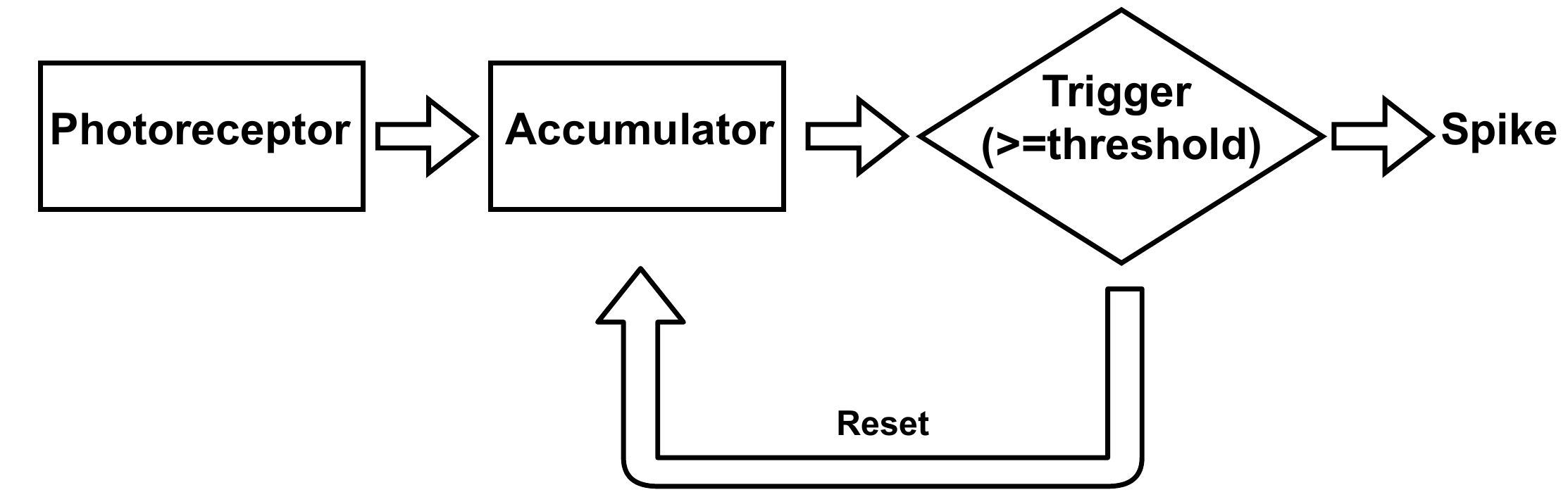}
\centering
\caption{Fovea-like sampling workflow}\label{fig2}
\end{figure}
    Specifically, in Spike camera, the intensity of light is converted into voltage by the photoreceptor. Once the analog-to-digital converter (ADC) completes the signal conversion and outputs the digital brightness, the accumulator at each pixel accumulates the brightness. At the moment \begin{small}$t$\end{small}, for pixel \begin{small}$(i, j)$\end{small}, if the accumulated brightness arrives a fixed threshold \begin{small}$\phi$\end{small} (as (1)), then a spike is fired and the corresponding accumulator is reset.
   \begin{eqnarray}
    B(i, j, t) = \int_{t_{i, j}^{pre}}^{t} I(i, j, \tau) d\tau\geq \phi, 
    \end{eqnarray}
    where \begin{small}$B(i, j, t)$\end{small} is the accumulated brightness at sampling time \begin{small}$t$\end{small}, \begin{small}$I(i, j, \tau)$\end{small} refers to the brightness of pixel \begin{small}$(i, j)$ \end{small}  at time \begin{small}$\tau$\end{small}, and \begin{small} $t_{i, j}^{pre}$\end{small} expresses the last time when spike is fired at pixel \begin{small}$(i, j)$ \end{small} before time \begin{small}$t$ \end{small}. If \begin{small} $t$\end{small} is the first time to send a spike, then \begin{small}$t_{i, j}^{pre}$ \end{small} is set 0.  Further, spike data can be mathematically defined as, 
   \begin{eqnarray}
    S_{FSM}(i, j, t) = \begin{cases} 
    1 &\mbox{ if (1) is satisfied}, \\
    0 &\mbox{ if (1) is not satisfied},  \\
    \end{cases} 
    \end{eqnarray}
     where, for pixel \begin{small}$(i, j)$ \end{small}, if a spike is outputted at sampling time \begin{small}$t$ \end{small}, \begin{small}$S(i, j, t)$ \end{small} is set digital signal “1”,  otherwise \begin{small} $S(i, j, t)$\end{small} is set “0”.
     Accordingly, the average brightness of pixel \begin{small}$(i, j)$ \end{small} between time \begin{small}$t$ \end{small} and \begin{small}$t_{i, j}^{pre}$ \end{small} can be calculated  \cite{spikecamera}, i.e., 
    \begin{eqnarray}
    \bar I(i, j)  \approx \dfrac{\phi}{t - t_{i, j}^{pre}} = \dfrac{\phi}{{n\Delta t}}, 
    \end{eqnarray}
    where, \begin{small}$\Delta t$ \end{small} is the sampling time interval and \begin{small}$n \in \mathbb{N}$ \end{small} denotes the number of intervals. 
    Hence, Spike camera can capture the visual texture of objects in all kinds of scenes including static and dynamic. 
\begin{figure}[t]
\includegraphics[width=8cm]{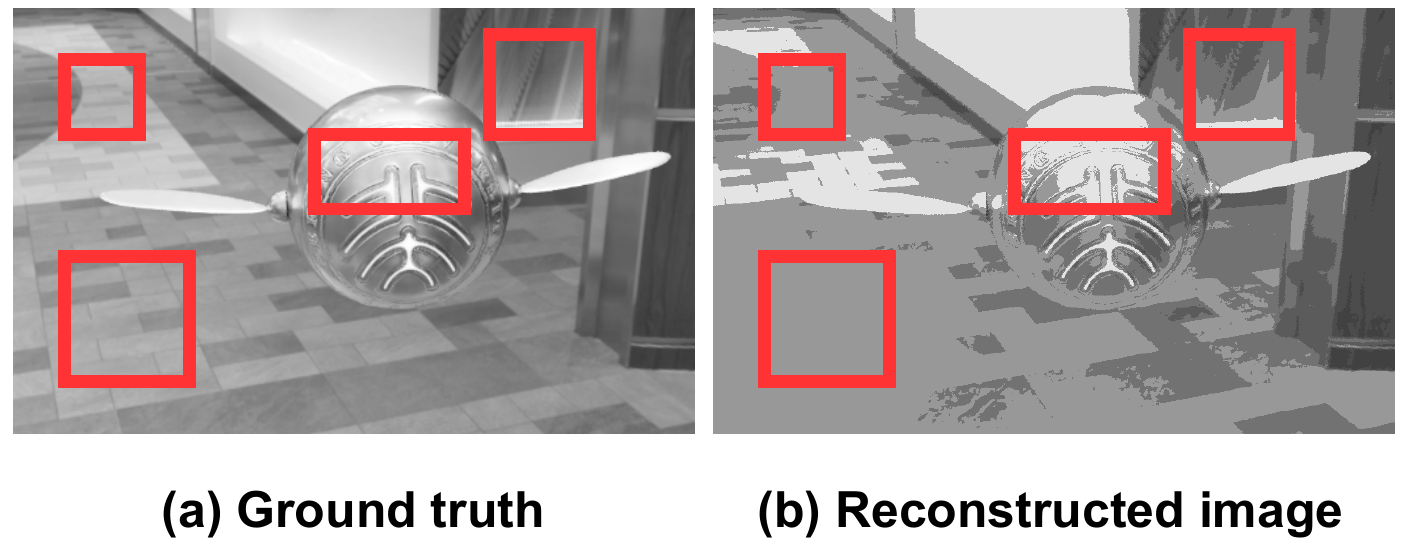}
\centering
\caption{Quantization error in Spike camera. (a) is a
virtual scene. (b) is the reconstructed image from spike data. The whole sampling process is simulated in a Spike camera simulator \cite{sim2}, threshold $\phi = 400$ and the reconstructed method is TFI \cite{spikecamera}. }\label{fig3}
\end{figure}
\begin{figure}[t]
\includegraphics[width=8cm]{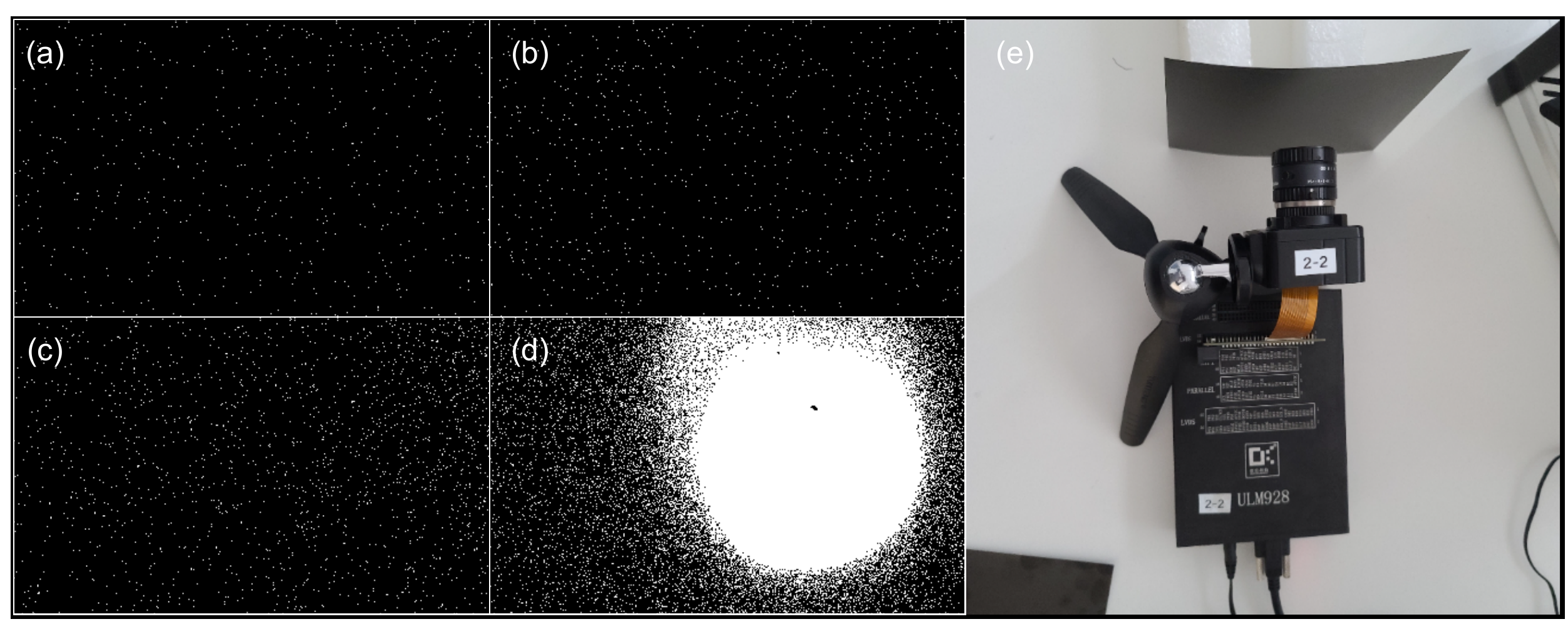}
\centering
\caption{Noise in Spike camera. We use the Spike camera to sample four kinds of spike data under different light conditions and the sampled scene is a black paper. (a)(b)(c)(d) are respectively from four kinds of spike data. (a) is sampled under no illumination in the room. (b) is sampled under weak illumination in the room. (c) is sampled under strong illumination in the room. (d) is sampled under direct illumination in the room. All white dot in (a)(b)(c)(d) is from noise and the number of white dot increases with the increase of illumination. (e) is our experimental device.}
\end{figure}
\subsection{Quantization Error}
     In fact, Spike camera in complex environments is not ideal enough, i.e., there is texture blur in reconstructed images (fig3). Especially for extreme light conditions, e.g., direct sunlight, the texture blur is more obvious. Quantization error is one of the main factors leading to above problem. Before that, quantization error in Spike camera has not been discussed in detail. In the work, we first give corresponding theoretical analysis. Quantization error comes from discrete sampling which results in spike is sent discretely and $\bar I(i, j)$ in a certain range can be estimated to be the same value. Although quantization error cannot be avoided, it can be reduced by increasing threshold. The cost of the method is to increase the response time of Spike camera to scenes, i.e., the time from the beginning of sampling to first spike. Hence, increasing threshold is not a good method to improve the performance of Spike camera. The proof of above conclusion is given in the appendix. In addition, \cite{recon1} uses network to post process spike data to improve the quality of reconstructed images. However, it can introduce a lot of extra time and space costs.
     
    
\subsection{Receptive Field Model}
FSM mimicking human visual system ignores some important mechanisms, e.g., receptive field mechanism of ganglion cells which causes the loss of some function in sampling. Specifically, FSM considers that a bipolar cell only connects one photoreceptor and one ganglion cell, but ganglion cells can actually respond to a field of photoreceptors referred to receptive field \cite{bio1, bio2, bio3, bio4, bio5} (as fig1). The receptive field mechanism plays an important role in describing image texture information distinctly and robustly \cite{derf}. Different ganglion cells have receptive fields of different scales and different polarity (on or off, as fig1(d)) which controls the size of response area and the sensitivity to light changes respectively \cite{recep1, recep2, recep3, recep4}. And Gaussian filter bank (as (4)) is a popular model for receptive field  \cite{freak, daisy}. However, the signal representation ability of Gaussian filter bank is limited which means the information in scene cannot be fully characterized. Beisdes, receptive field can also be simulated by DoG filter  \cite{bio1}. And  \cite{derf} use wavelet filter bank, generated by DoG filters consisting of the difference of two Gaussian filters, to simulate the receptive field of ganglion cells and making a big success on local image descriptor. 
\\\indent
Accordingly, the process where ganglion cells deal with the electrical signal from photoreceptors is similar to wavelet transform \cite{bio1}. After ganglion cells have processed electrical signal, they decide whether to fire a spike according to their own activation level \cite{bio4, bio5}.
Although wavelet transform is suitable to model the receptive field mechanism and its related theories are mature \cite{wave1, wave2, wave3}, the transform is not used in the visual sampling of Spike camera. Hence, how to embed wavelet transform into visual sampling process is a problem worthy of study.
\begin{figure}[htb]
\includegraphics[width=0.8\linewidth]{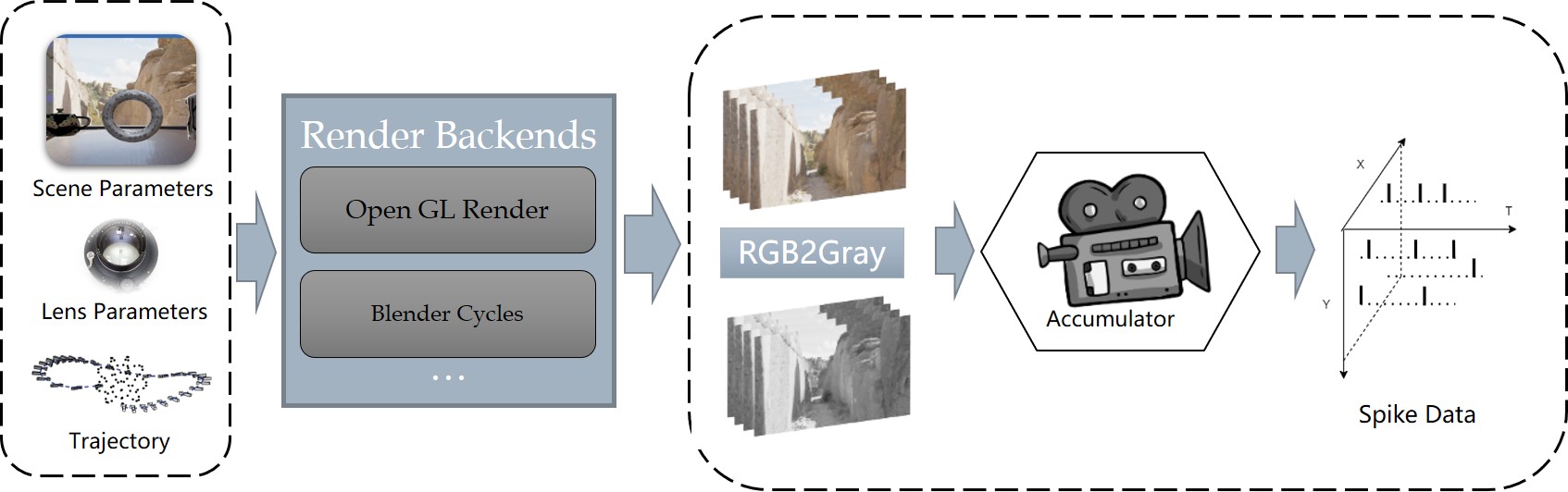}
\centering
\caption{The function ``v2s" in the Spike camera simulator where image sequences can be converted to spike data.}\label{fig4}
\end{figure}
\begin{figure*}[htbp]
\includegraphics[width=\linewidth]{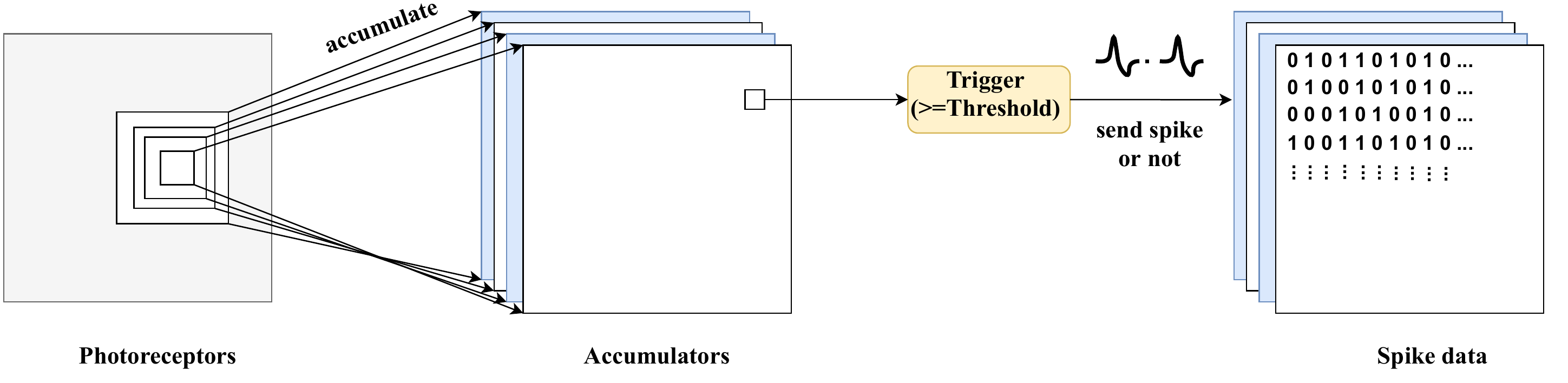}
\centering
\caption{The workflow of RVSM. Firstly, every accumulator accumulates the brightness in Photoreceptos according to own receptive field. Then, the accumulators completing trigger condition fire spike and reset own accumulation. And spike data records whether accumulators send spike or not each sampling.}\label{fig5}
\end{figure*}
\subsection{Spike Camera Simulator}
All our experiments are finished in the Spike camera simulator \cite{sim2}. The Spike camera simulator provides the simulation only for Spike camera and has some same extended function as esim \cite{sim1} (a simulator for the sampling of DVS).
Specifically, the Spike camera simulator implements an approximate simulation about the sampling mechanism of the Spike camera in the time domain and the space domain. We can introduce RVSM into the Spike camera simulator and use it to convert videos to spike data (as fig5).
\section{A Robust Visual Sampling Model Inspired by Receptive Field}
\subsection{Model Architecture}
RVSM is a bio-inspired visual sampling model which can capture the texture of objects in all kinds of scenes. Different from FSM, RVSM considers receptive field mechanism in the human visual system and is more similar to biological visual sampling. And, in RVSM, the intensity of light is converted into voltage by the photoreceptor. Once the analog-to-digital converter (ADC) completes the signal conversion and outputs the digital luminance intensity, the accumulator at each pixel can accumulate the weighted sum of intensity in its own receptive field where weighted is controlled by used filter, called as "summation process". Here, we use the normalized DoG filter to simulate the receptive field (RVSM$_{DoG}$) and it can be generated by Gaussian filter (as (4)).
\begin{equation}
G_{\sigma}^{x_0, y_0}(i, j) = \dfrac{1}{{2*\pi}\sigma^2}\exp^{-\dfrac{(i - x_0)^2 + (j - y_0)^2}{2\sigma^2}}, 
\end{equation}
where \begin{small}$G_{\sigma}^{x_0, y_0}(i, j)$\end{small} is Gaussian filter, \begin{small}$\sigma$ \end{small} is standard deviation and \begin{small}$(x_0, y_0)$ \end{small} is expectation. Accordingly, the DoG filter (as (5)) can be as the mother wavelet of  normalized DoG filter bank.
    \begin{equation}
    \begin{split}
    DoG(i, j)= {G_{a_1}^{0, 0}(i, j) - G_{a_2}^{0, 0}(i, j) }, 
    \end{split}
    \end{equation}
where, we set \begin{small}$a_1 = 1$\end{small},\begin{small}$a_2 = 1.5874$\end{small}. Further, we can get normalised DoG filter bank by translation, scaling and normalization,
    \begin{equation}
    \begin{split}
    DoG_{\sigma}^{x_0, y_0}(i, j)= \dfrac{DoG(\dfrac{i-x_0}{\sigma},\dfrac{j-y_0}{\sigma})}{\sum\limits_{(p, q)\in C_{\sigma}^{x_0, y_0}} \!\!\!\!\!\!\!\! |DoG(\dfrac{p-x_0}{\sigma},\dfrac{q-y_0}{\sigma})|},
    \end{split}
    \end{equation}
where \begin{small}$i, j \in \mathbb{Z}$\end{small},  \begin{small} $C_{\sigma}^{x_0, y_0} = [x_0-L_{\sigma}, x_0+L_{\sigma}] \times[y_0-L_{\sigma}, y_0+L_{\sigma}] \cap \mathbb{Z}^2$\end{small}, \begin{small}$L_{\sigma} \in \mathbb{Z}$\end{small} is the template size decided by the scale of receptive field, \begin{small} $\sigma$\end{small} controls the scale of receptive field. And the above summation process can be expressed as,
   \begin{eqnarray}
    A_{\sigma}^{x_0, y_0}(t) =\!\!\!\!\!\!\! \sum\limits_{(i, j)\in C_{\sigma}^{x_0, y_0}} \!\!\!\!\!\!\!DoG_{\sigma}^{x_0, y_0}(i, j) \int_{t_{DoG_{\sigma}^{x_0, y_0}}^{pre}}^{t} \!\!\!\!\!\!\!\!\!\!\!\!\!\!\! I(i, j, \tau) d\tau, 
    \end{eqnarray}
where \begin{small}$A_{\sigma}^{x_0, y_0}(t)$\end{small} expresses the accumulation of an accumulator with receptive field \begin{small}$DoG_{\sigma}^{x_0, y_0}$\end{small} in pixel \begin{small}$(x_0, y_0)$\end{small} at sampling time \begin{small}$t$\end{small},  \begin{small}$t^{pre}_{DoG_{\sigma}^{x_0, y_0}}$\end{small} is the last time when a spike is fired by the normalized DoG filter \begin{small}$DoG_{\sigma}^{x_0, y_0}$\end{small} before sampling time \begin{small}$t$\end{small} and is initially set to 0, and the number of normalised DoG filter is limited, i.e., \begin{small}$\sigma$\end{small} take finite values. And we assume the set of all possible \begin{small}$\sigma$\end{small} is \begin{small}$P$\end{small}. In particular, RVSM$_{DoG}$ is the same as FSM when we only use a normalized DoG filter with unit scale (the template size of filter is 1) to sample. If the absolute value of accumulation of an accumulator arrives a fixed threshold \begin{small}$\phi$\end{small}, a spike can be fired. Hence, the spike data from RVSM$_{DoG}$ can be expressed as, 
   \begin{eqnarray}
    S_{\sigma}(x_0, y_0, t) = 
    \begin{cases} 
    1 &\mbox{ if $A_{\sigma}^{x_0, y_0}(t) >= \phi$}, \\
    -1 &\mbox{ if $ A_{\sigma}^{x_0, y_0}(t) <= -\phi$},  \\
    0 &\mbox{ else}, \\
    \end{cases} 
    \end{eqnarray}
where \begin{small}$\phi \geq 0$\end{small}, for pixel \begin{small}$(x_0, y_0)$\end{small}, \begin{small}$S_{\sigma}(x_0, y_0, t)$\end{small} is set digital signal “1” (“-1”) if corresponding accumulation arrives threshold (negative threshold), otherwise \begin{small}$S_{\sigma}(x_0, y_0, t)$\end{small} is set “0”. 
After a spike is fired, the accumulation of corresponding accumulator is reset. Beisdes, we can also use normalized Gaussian filter to simulate the receptive field in the sampling model (we called RVSM$_{Gauss}$), i.e., replacing normalized DoG filter with normalized Gaussian filter. Detailed formula is in appendix. The whole sampling workflow is shown in fig6.
\subsection{Visual Texture Reconstruction}
\begin{figure*}[htbp]
\includegraphics[width=\linewidth]{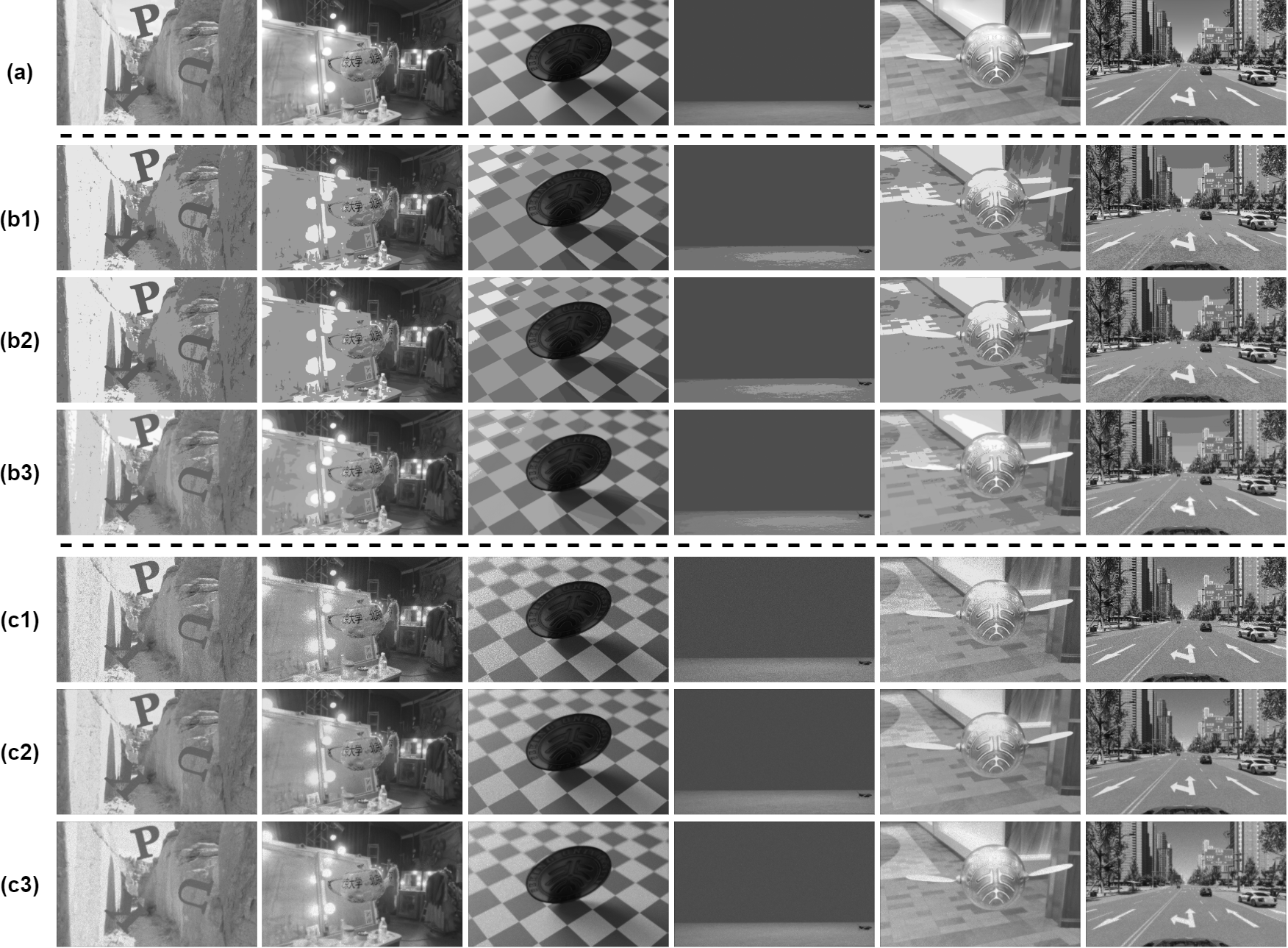}
\centering
\caption{The result of reconstructed images. From left to right, scenes are ``Character", ``Teapot", ``Coin", ``Grasshopper", ``Flyball" and ``Driving" respectively. Besides, (a) is ground truth, the reconstructed images in (b1)(b2)(b3) corresponds to spike data sampled by FSM, RVSM$_{Gauss}$ (One Gauss) and RVSM$_{DoG}$ (Three DoG) respectively in the absence of noise and the reconstructed images in (c1)(c2)(c3) corresponds to spike data sampled by FSM, RVSM$_{Gauss}$ (Four Gauss) and RVSM$_{DoG}$ (Four DoG) respectively in the presence of noise where the noise intensity uses default settings in Spike camera simulator.}\label{fig6}
\end{figure*}
\begin{table*}[ht]\large  
  \centering  
  \begin{spacing}{1.1}  
  \renewcommand\tabcolsep{10.0pt} 
  \resizebox{\hsize}{!}{  
    \begin{tabular}{c|c|cccccccccccc} 
    \toprule[2pt]
    \hline
    \multirow{2}*{\textbf{Metric}} & \multirow{2}*{\textbf{Method}} & \multicolumn{12}{|c}{\textbf{Scene}}
    \\
    \cline{3 - 14}
     & & \textbf{Character} & \textbf{Character(N)} & \textbf{Teapot} & \textbf{Teapot(N)} & \textbf{Coin} & \textbf{Coin(N)}
     & \textbf{Grasshopper} & \textbf{Grasshopper(N)}
     & \textbf{Flyball} & \textbf{Flyball(N)}
     & \textbf{Driving} & \textbf{Driving(N)}
     \\
    \hline
    \multirow{9}*{\textbf{MSE}} & FSM & 129.29 & 400.43 & 187.38 & 318.33 & 352.71 & 488.72
 & 60.78 & 123.58 & 432.64 & 653.92 & 186.32 & 316.69
 \\
     &  One DoG & 170.44 & 653.81 & 216.27 & 505.21 & 381.64 & 774.12 & 68.27 & 199.09 &477.32 &1031.74 &297.91 &557.42
 \\
     &  Two DoG & 152.06 & 586.81 & 206.24 & 455.25 & 370.52 & 686.57 & 54.38 & 166.22 &459.16 &927.13 &344.85 &568.38
 \\
     &  Three DoG & \textbf{96.16} & 283.18 & \textbf{87.47} & 204.87 & 135.59 & 289.90 & 56.65  & 105.88 &150.74 &373.40 &\textbf{160.69} &264.39
\\
     &  Four DoG & 109.86 & 209.19 & 93.34 & 130.23 & \textbf{111.12} & 132.19 & \textbf{51.46} & 60.32 &\textbf{127.85} & 182.14&217.25 &242.17
 \\
     &  One Gauss & 119.93 & 320.79 & 182.46 & 262.76 & 355.08 & 398.87 & 59.26 & 100.40 &431.44 &533.63 &185.38 &271.96
 \\
     &  Two Gauss & 110.56 & 139.77 & 175.69 & 129.68 & 351.99 & 185.62 & 57.59 & 46.84 &420.70 &249.24 & 196.18&\textbf{163.05}
 \\
     &  Three Gauss & 109.50 & 98.24 & 174.79 & 100.73 & 353.63 & 133.90 & 57.67 & 33.09&418.12 &179.57 &232.50 &172.86
 \\
     &  Four Gauss & 112.74 & \textbf{89.02} & 176.92 & \textbf{95.83} & 357.77 & \textbf{118.84} & 58.55 & \textbf{28.67} &419.15 & \textbf{157.89}&277.38 &209.13
 \\
    \hline
    \hline
    \multirow{9}*{\textbf{PSNR}} & FSM & 27.02 & 22.11 & 25.41 & 23.10 & 22.68 & 21.26 & 30.29 & 27.21 &21.86 &20.02 &25.43 &23.12
 \\
     &  One DoG & 25.81 & 19.97 & 24.78 & 21.09 & 22.33 & 19.25 & 29.79 & 25.14 &21.42 &18.04 &23.38 &20.67
 \\
     &  Two DoG & 26.31 & 20.45 & 24.98 & 21.55 & 22.47 & 19.78 & 30.77 & 25.92 &21.59 &18.50 &22.75 &20.59
 \\
     &  Three DoG & \textbf{28.30} & 23.61 & \textbf{28.71} & 25.01 & 26.83 & 23.52 & 30.59 & 27.88 &26.37 &22.43 & \textbf{26.07}&23.91
 \\
     &  Four DoG & 27.72 & 24.92 & 28.43 & 26.98 & \textbf{27.75} & 26.96 & \textbf{31.06} & 30.32 &\textbf{27.12} &25.53 &24.76 &24.28
 \\
     &  One Gauss & 27.34 & 23.06 & 25.51 & 25.51 & 22.65 & 22.14 & 30.40 & 28.11 & 21.88&20.91 & 25.45&23.78
 \\
     &  Two Gauss & 27.69 & 26.67 & 25.68 & 27.00 & 22.69 & 25.46 & 30.52 & 31.42 & 21.99 &24.21 & 25.20&\textbf{26.01}
 \\
     &  Three Gauss & 27.73 & 28.19 & 25.70 & 28.09 & 22.67 & 26.88 & 30.52 & 32.93&22.02 &25.63 & 24.47&25.75
 \\
     &  Four Gauss & 27.61 & \textbf{28.63} & 25.65 & \textbf{28.31} & 22.62 & \textbf{27.41} & 30.45 & \textbf{33.55} &22.00 &\textbf{26.19} &23.70 &24.92
 \\
    \hline
    \hline
    \multirow{9}*{\textbf{SSIM}} & FSM & 0.749 & 0.346 & 0.837 & 0.524 & 0.881 & 0.362 & 0.943 &0.438 &0.770 & 0.266 &0.801 &0.581
 \\
     &  One DoG & 0.707 & 0.267 & 0.812 & 0.434 & 0.854 & 0.288 & 0.930 & 0.318 & 0.747& 0.203 & 0.748&0.522
 \\
     &  Two DoG & 0.724 & 0.285 & 0.824 & 0.455 & 0.857 & 0.306 & 0.931 & 0.352 &0.757 &0.214 & 0.739&0.523
 \\
     &  Three DoG & 0.797 & 0.413 & 0.876 & 0.586 & 0.901 & 0.432 & 0.947 & 0.539 &0.811 &0.319 & 0.821&0.622
 \\
     &  Four DoG & \textbf{0.803} & 0.535 & \textbf{0.888} & 0.703 & \textbf{0.919} & 0.626 & \textbf{0.965} & 0.752 &\textbf{0.852} &0.494 & \textbf{0.854}&0.697
 \\
     &  One Gauss & 0.760 & 0.387 & 0.842 & 0.563 & 0.889 & 0.396 & 0.946 & 0.495 &0.778 &0.298 &0.808 &0.598
 \\
     &  Two Gauss & 0.777 & 0.582 & 0.851 & 0.719 & 0.893 & 0.561 & 0.952 & 0.712 & 0.786&0.466 &0.816 &0.703
 \\
     &  Three Gauss & 0.787 & 0.694 & 0.854 & 0.798 & 0.895 & 0.669 & 0.955 & 0.821 &0.791 &0.588 & 0.806&0.749
 \\
     &  Four Gauss & 0.789 & \textbf{0.755} & 0.854 & \textbf{0.840} & 0.894 & \textbf{0.739} & 0.957 & \textbf{0.878} & 0.793& \textbf{0.671} &0.786 &\textbf{0.760}
 \\
    \hline
    \bottomrule[2pt]
    \end{tabular}
    }
  \end{spacing}
    \caption{The quantitative metrics on HMD where the spike data is sampled in the presence of noise for Scene(N) and the spike data is sampled without noise for Scene.} 
  \label{table5} 
\end{table*}
Spike data from FSM can be easily used to reconstruct scene information according to its actual meaning, i.e., a spike means brightness accumulation is large enough. Similarly, to illustrate the validity of spike data from RVSM, we also provide an easy method to restore the captured scene according to actual meaning of spike data from RVSM. In RVSM, a spike means that the weighted accumulation of brightness in receptive field arrives activation level. Further, a spike in RVSM$_{DoG}$ means the absolute value of the coefficient of the brightness accumulation signal on a normalized DoG basis is large enough because the whole summation process is realized by an inner product operation (as (7)). Hence, spike data from RVSM$_{DoG}$ can report brightness accumulation signal in wavelet domain. We assume the coefficient matrix of brightness accumulation signal in wavelet domain is \begin{small}$K_{\sigma}^{i, j}(t)$\end{small}. And \begin{small}$K_{\sigma}^{i, j}(t)$\end{small} can be estimated as, 
    \begin{align}
    K_{\sigma}^{i, j} & (t) \approx
    \\ &
    \begin{cases} 
    \dfrac{\phi}{t -  t_{DoG_{\sigma}^{i, j}}^{pre}} &\mbox{ if $S_{\sigma}(x_0, y_0, t) = 1$}, \\\\
    \dfrac{-\phi}{t -  t_{DoG_{\sigma}^{i, j}}^{pre}} &\mbox{ if $S_{\sigma}(x_0, y_0, t) = -1$},  \\\\
    K_{\sigma}^{i, j}(t - 1) &\mbox{ else}, \\ \nonumber
    \end{cases} 
    \end{align}

where \begin{small}${t -  t_{DoG_{\sigma}^{i, j}}^{pre}}$\end{small} expresses the time from the last spike to present, \begin{small}$K_{\sigma}^{i, j}(t)$\end{small} can be updated if the accumulator with receptive field \begin{small}$DoG_{\sigma}^{i, j}$\end{small} fires a spike at the sampling time \begin{small}$t$\end{small}, otherwise \begin{small}$K_{\sigma}^{i, j}(t)$\end{small} uses the coefficient  at the sampling time \begin{small}$t - 1$\end{small}, and \begin{small}$K_{\sigma}^{i, j}(0)$\end{small} is set 0.
Accordingly, we can obtain the approximate brightness accumulation signal using the inverse wavelet transformation of the coefficient matrix,
\begin{eqnarray}
I(x, y, t) \approx \sum\limits_{i, j, \sigma} K_{\sigma}^{i, j}(t) DoG_{\sigma}^{i, j}(x, y)
\end{eqnarray}
where \begin{small}$I(x, y, t)$\end{small} denotes the brightness of pixel \begin{small}$(x, y)$\end{small} at sampling time \begin{small}$t$\end{small},  \begin{small}$(i, j) \in \mathbb{Z}^2 \cap ([1, height] \times [1, width])$\end{small}, \begin{small}$height \times width$\end{small}  controls the resolution size of sampling and \begin{small}$\sigma \in P$\end{small}. Similarly, spike data from RVSM$_{Gauss}$ and more general RVSM also can estimate brightness of scenes and detailed formula is in appendix.
\subsection{The Generalization of RVSM}
Although, we choose Spike camera as the carrier of RVSM due to great potential of Spike camera, this does not mean that RVSM is only suitable for Spike camera. As an idea inspired by receptive field to collect regional information, RVSM can also be used in other neuromorphic vision sensors e.g., DVS and we just need to make corresponding changes according to the principle of different sensors. The related detail is in appendix.
\section{Experiment}
\subsection{Dataset}
To fully compare the sampling performance among RVSM$_{DoG}$, RVSM$_{Gauss}$ and FSM, we provide a high-speed motion spike dataset (HMD) including six scenes. The dataset has $6 \times 9 \times 2$ spike sequences, i.e., each scene contains $9\times 2$ spike sequences (with or without noise) captured by FSM, RVSM$_{DoG}$ with four kinds of normalized DoG filter bank (we called them as One DoG, Two DoG, Three DoG and Four DoG, the corresponding \begin{small}$P$\end{small} is \begin{small}$\{0.24\}, \{0.24,\; 0.348\}, \{0.24,\; 0.348,\; 0.5046\}, \{0.24,\; 0.348, \\ 0.5046,\, 0.7317\}$\end{small} and RVSM$_{Gauss}$ with four kinds of normalized Gaussian filter bank (we called them as One Gauss, Two Gauss, Three Gauss and Four Gauss) respectively. The configuration of noise is the same as simulator. Note that, for fairness, we ensure that RVSM$_{DoG}$ and RVSM$_{Gauss}$ have the same response time to scenes with FSM and the related details are in appendix. The above spike data is generated by Spike camera simulator. Besides, the dataset also has 6 image sequences as ground truth. The six class scenes are named as ``Character", ``Teapot", ``Coin", ``Grasshopper", ``Flyball" and ``Driving" where ``Character" corresponding to 500 images describes the characters with simple rotation (uniform rotation on a dimension), ``Teapot" corresponding to 500 images describes a teapot with easy rotation (uniform rotation in three dimensions), ``Coin" corresponding to 500 images describes the complex rotation of a coin, ``Grasshopper" corresponding to 500 images describes a simple motion of grasshopper jumping, ``Flyball" corresponding to 500 images describes a complex motion of flying ball and ``Driving" corresponding to 500 images describes vehicle driving in complex scenes.
\subsection{The Performance of RVSM}
We compare the sampling performance among RVSM$_{DoG}$, RVSM$_{Gauss}$ and FSM by calculating the metric (PSNR, MSE and SSIM) of reconstructed images from RVSM$_{DoG}$ (One DoG, Two DoG, Three DoG and Four DoG), RVSM$_{Gauss}$ (One Gauss, Two Gauss, Three Gauss and Four Gauss) and FSM respectively. For spike data from FSM, we use TFI \cite{spikecamera} to reconstruct images because TFP \cite{spikecamera} is not suitable for spike data sampled in high-speed scenes. For One DoG, Two DoG, Three DoG and Four DoG (One Gauss, Two Gauss, Three Gauss and Four Gauss), the latter introduces a normalized DoG filter (normalized Gaussian filter) with a larger scale than the former and their the minimum scale of filter corresponds to template size 3x3. Besides, the reconstructed images are adjusted to the same brightness level as ground truth. 
\\\indent Fig7 shows the experimental results. In the absence of noise (fig7(b1)(b2)(b3)), we can find that, for complex scenes e.g., ``Flyball", RVSM$_{DoG}$ can capture fine textures and has a higher contrast. From table 1, we can also get a consistent conclusion that the images from RVSM$_{DoG}$ (Three DoG and Four DoG) have much better quality than the images from FSM and RVSM$_{Gauss}$ in the absence of noise. It shows that RVSM$_{DoG}$ is so less affected by quantization error that the model can more effectively sample the texture information in high-speed motion scenes. Although the performance of RVSM$_{Gauss}$ is far inferior to that of RVSM$_{DoG}$, RVSM$_{Gauss}$ (One Gauss) also has similar performance in most scenes to FSM. The above results shows that RVSM by introducing receptive field mechanism to sample regional information is an effective visual sampling model for Spike camera. And how to choose the filter bank used to simulate receptive field is so important which decides the performance of RVSM.
\\\indent
RVSM still shows powerful performance compared with FSM in the presence of noise (as Table 1). The conclusion is a little different from that without noise. First, for all scenes, the quality of reconstructed images from RVSM$_{DoG}$ (Four DoG) and RVSM$_{Gauss}$ (Four Gauss) is much better than that from FSM. It means that, in a more real environment (with noise), RVSM can sample the information of scenes more effectively due to suffering less quantization error and noise error and improves the problem that Spike camera is sensitive to noise largely. Interestingly, by comparing the change of quantization metrics before and after adding noise, we also find that the reconstruction result from RVSM$_{Gauss}$ is better in the presence of noise than in the absence of noise because noise offsets part of the quantization error. Besides, as more the receptive field with large scale is used to sample (from One DoG to Four DoG, from One Gauss to Four Gauss), the quality of reconstructed images is greatly improved for most scenes. This is because the receptive field with large scale is more robust to noise and the conclusion is confirmed by the subsequent robustness experiments. 
\subsection{The Robustness of RVSM}
\begin{figure*}[ht]
\includegraphics[width=\linewidth]{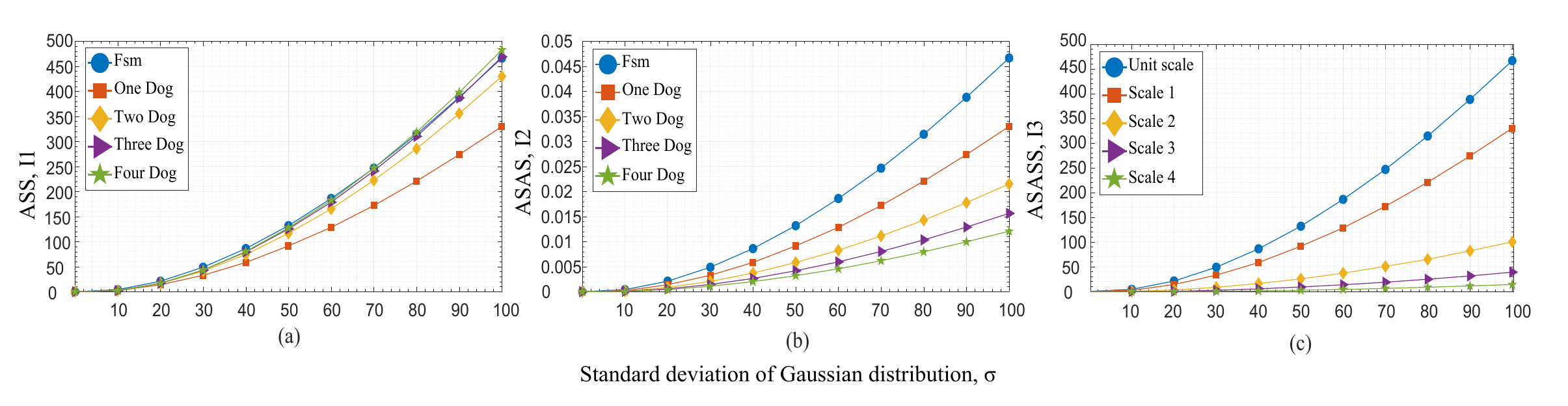}
\centering
\caption{The influence of noise intensity on ASS $I_1$, ASAS $I_2$ and ASASS $I_3$ where the scale set $\sigma$ of Scale 1, Scale 2, Scale 3 and Scale 4 is 0.24, 0.348, 0.5046 and 0.7317.}\label{fig8}
\end{figure*}
In the actual sampling process, noise is nowhere and excellent sampling methods can effectively filter all kinds of noise. Hence, we study the effects of different intensities of noise on RVSM$_{DoG}$ and FSM. In FSM and RVSM$_{DoG}$, Noise mainly occurs in the process of light intensity accumulation  \cite{sim2}. Here, we consider the noise caused by dark electric current, the offset voltage and capacitor. Further, the process of light intensity accumulation and spike data for RVSM$_{DoG}$ can be updated as,
   \begin{align}
    \!\!\!\!A_{\sigma}^{x_0, y_0}(t) & =\!\!\!\!\!\! \sum\limits_{(i, j)\in C_{\sigma}^{x_0, y_0}} \!\!\!\!\!\!DoG_{\sigma}^{x_0, y_0}(i, j) \!\!\int_{t_{DoG_{\sigma}^{x_0, y_0}}^{pre}}^{t} \!\!\!\!\!\!\!\!\!(I(i, j, \tau) \nonumber
    \\ \\
    & + I_{dark}(i, j, \tau)) d\tau, \nonumber
    \end{align}
   \begin{align}
    S_{\sigma}&(x_0, y_0, t) =
    \\ &
    \begin{cases} 
    1 &\mbox{ if \begin{small} $A_{\sigma}^{x_0, y_0}(t) >= \theta(i, j) \phi + V_{OS}(i, j)$\end{small}}, \\
    -1 &\mbox{ if \begin{small}$ A_{\sigma}^{x_0, y_0}(t) <= -(\theta(i, j) \phi + V_{OS}(i, j))$\end{small}},  \\
    0 &\mbox{ else}, \nonumber
    \end{cases} 
    \end{align}
where $I_{dark}(i, j, \tau)$, $V_{OS}(i, j)$ and $\theta(i, j)$ are noise random variable. They denote the dark electric current, the offset voltage and capacitor noise respectively. For FSM, the process of light intensity accumulation and spike data is in simulator.
\\\indent
We design an easy scene to test the robustness of the sampling model in Spike camera simulator. In the scene, the background is black which means that the work current $I$ is ``0". Therefore, all spikes are generated by the noise. Further, we use three kinds of the index to describe the robustness of sampling models, i.e., the average number of spike per sampling (ASS, $I_1$), the average number of spike generated by each accumulator per sampling (ASAS, $I_2$) and the average number of spike generated by all accumulators with the same scale per sampling (ASASS, $I_3$). The ASS can be defined as,
\begin{equation}
I_1 = \begin{cases} 
\dfrac{\sum\limits_{i,j,t} S_{FSM}(i, j, t)}{T} &\mbox{ for FSM }, \\\\
\dfrac{\sum\limits_{\sigma,i,j,t} |S_{\sigma}(i, j, t)|}{T} &\mbox{ for RVSM$_{DoG}$},  \\
\end{cases}
\end{equation}
where \begin{small}$T$\end{small} is set to 1000 which denotes the total number of sampling, \begin{small}$\sigma \in P$\end{small}, \begin{small}$(i, j) \in \mathbb{Z}^2 \cap [1, H] \times [1, W]$\end{small}, \begin{small}$H \times W$\end{small}  controls the resolution size of sampling and here \begin{small}$H$\end{small} and \begin{small}$W$\end{small} are both set to 100. Further, the ASAS can be expressed as,
\begin{equation}
I_2 = \begin{cases} 
\dfrac{I_1}{WH} &\mbox{ for FSM }, \\\\
\dfrac{I_1}{WH|P|} &\mbox{ for RVSM$_{DoG}$},  \\
\end{cases}
\end{equation}
where $|P|$ is the number of elements in $P$. And the ASASS can be expressed as, (\begin{small}$\sigma$\end{small}) as,  
\begin{equation}
I_3(\sigma) = \begin{cases} 
I_1 &\mbox{ for FSM }, \\\\
\dfrac{\sum\limits_{i,j,t}|S_{\sigma}(i, j, t)|}{WH} &\mbox{ for RVSM$_{DoG}$},  \\
\end{cases}
\end{equation}
Besides, for simplicity, we assume that the noise are independent and identically distributed. In the experiment, \begin{small}$I_{dark}(i, j, \tau)$\end{small}, \begin{small}$V_{OS}(i, j)$\end{small} and \begin{small}$\theta(i, j)$\end{small} are set to Gaussian distribution, i.e., 
\begin{align}
I_{dark}(i, j, \tau) \sim N(e_1, (\beta_1 *k)^2), \\
V_{OS}(i, j) \sim N(e_2, (\beta_2 *k)^2), \\
\theta(i, j) \sim N(e_3, (\beta_3 *k)^2), 
\end{align}
where $e_1$, $e_2$ and $e_3$ are expectation, $\beta_1*k$, $\beta_2*k$ and $\beta_3*k$ are standard deviation and we use $k$ to control their noise intensity.
\\\indent
The result is showed in fig8. In fig8(a), we can find the average number of spike per sampling increases with the increase of standard deviation, i.e., the amount of noise increases. Besides, the average number of spike per sampling in RVSM$_{DoG}$ is less when the standard deviation is lower than some fixed value and the average number of spike per sampling in Three DoG and Four DoG is more than FSM a little when the standard deviation is large. It means RVSM$_{DoG}$ can produce less noise than FSM when the noise intensity is not large. fig8(b) shows the average number of spike generated by each accumulator per sampling is less than FSM for all standard deviation. It means accumulators in RVSM$_{DoG}$ have a better ability to filter noise. Further, we can see that, with the introduction of accumulators with a larger scale, the average number of spike generated by each accumulator per sampling decreases. Hence, the accumulator with large scale is more resistant to noise. And this conclusion is more directly verified by fig8(c), i.e., the average number of spike generated by accumulators with more large scale per sampling is less under the same standard deviation. Besides, RVSM$_{Gauss}$ has the similar robustness to RVSM$_{DoG}$ and, in appendix, we give a theoretical explanation about the robustness of RVSM in simple case. 

\section{Conclusion}
In this paper, we propose a novel sampling model (RVSM) for Spike camera which respectively uses wavelet filter bank (RVSM$_{DoG}$) and Gaussian filter bank (RVSM$_{Gauss}$) to simulate receptive field and is closer to the human visual system than FSM. The spike data from RVSM can report the brightness accumulation signal in function domain. Accordingly, we propose an efficient method similar to inverse wavelet transform to convert spike data from RVSM into images. Besides, we test the performance of FSM, RVSM$_{DoG}$ and RVSM$_{Gauss}$ in proposed HDM which is a high-speed motion spike dataset including a variety of motion scenes. Interestingly, we find the sampling with the receptive field (RVSM$_{DoG}$ and RVSM$_{Gauss}$) has the better ability to capture the texture information of objects than FSM. Further, we discuss the robustness of RVSM and FSM to noise. The results show that FSM suffers the attack of noise easily, while RVSM can filter high intensity noise effectively by collecting regional information. Besides, RVSM is not only suitable for Spike camera, but also for other neuromorphic vision sensors e.g., DVS. Next, we will study the performance of RVSM in other sensors and port it to hardware. All code can be released after the paper is published. 

\hupar{Acknowledgments.}This work is supported by grants from the National Natural Science Foundation of China under contract No.61806010.
\bibliographystyle{ieee_fullname}
\bibliography{egbib}
\end{document}